\documentclass[11pt]{article}

\title{Some medical applications of\\ Example-based super-resolution}
\author{Ramin Zabih}

\begin{document}

\maketitle

\begin{abstract}
Example-based super-resolution (EBSR) \cite{Freeman:CGA02,Freeman:IJCV00} 
reconstructs a high-resolution image from a low-resolution
image, given a training set of high-resolution images. In this note I propose some
applications of EBSR to medical imaging. A particular interesting application,
which I call \textit{x-ray voxelization}, approximates the result of a CT scan
from an x-ray image.
\end{abstract}

\section{Example-based super-resolution}

Example-based super-resolution (EBSR) \cite{Freeman:CGA02,Freeman:IJCV00}
produces a high-resolution image from a low resolution image, given a training
database of high-resolution images.  The basic idea is that an observed
low-resolution pixel is the average of a set of high-resolution pixels. The
training set can be viewed as a probabilistic map from a set of underlying
high-resolution pixels to the corresponding observed low-resolution pixel,
since in general it is straightforward to model the blurring and subsampling
process that turns a high-resolution image into a low-resolution one. Naively
EBSR could be solved using a nearest neighbors scheme, but of course it is
necessary to make use of inference and machine learning techniques to handle
the spatial relationships between pixels (see
\cite{Freeman:CGA02,Freeman:IJCV00} for details).

\section{Applications to medical imaging}

While super-resolution has many application, EBSR for medical applications
seems to have been under-explored (see \cite{Trinh:TMI14} for a rare
example). Yet there are many medical applications where EBSR would be
extremely natural.

As one example, consider CT scans, where perhaps the most serious limitation
is the risk posed by ionizing radiation \cite{Brenner:NEJM07}. For screening
purposes, lung CT is often done at low-dose, with the resulting loss of image
detail. Example-based super-resolution has the potential to provide the level
of detail that occurs with normal levels of radiation, while keeping the
radiation risk advantages of low-dose.

This would be achieved by collecting a database of normal-dose CT's along with
the corresponding low-dose images. If there are accurate simulations of the
noise and artifacts introduced by low-dose, it would be possible to model a
corresponding low-dose image from a normal-dose one. It is also possible that
under certain circumstances both a low-dose and a normal-dose image of a
patient may be available (for example, if the patient undergoes a low-dose
screening CT, and the findings warrant a follow-up normal-dose CT).

In this particular example, the problem is not precisely EBSR since the
low-dose and the high-dose images are likely to have identical resolution;
however, it is basically the same problem since the goal is to add details
using a database of training images. A similar application would involve
inferring the images seen in contrast-enhanced CT from non-contrast images,
since the IV contrast agents used have risks, especially for patients with
compromised kidney function.

Other natural medical examples appear in MRI, where the chief limitation is
scanning time. A collection of high-resolution images may be obtained as the
database, despite the extended scanning time, and used to fill in details for
low-resolution images obtained with faster scans. This could be particularly
useful for parts of the body that undergo motion. Obviously, some of these
applications would require registration as well (generally non-rigid
registration). 

\subsection{Advantages of medical applications for EBSR}

Medical applications of EBSR have some interesting advantages over conventional
photography (i.e., the application for which EBSR was developed).  

Spatial coordinates in medical applications are generally meaningful, while
they are of very limited utility in conventional photography.  For example, in
a chest CT pixels near the center of the image are very likely to be of the
heart or lungs, while pixels at the border are generally air.  In contrast,
any statistical argument about the content of pixels at the center of a
conventional photograph is likely to be extremely generic, and thus of limited
utility.  This holds out the promise that medical EBSR applications would
obtain higher resolution from a smaller dataset than most EBSR
applications. (Alternatively, medical EBSR could obtain better accuracy at the
same resolution.)

Related to this advantage is the fact that in medical imaging the size of a
pixel (or voxel) is specified in physical units (e.g., a voxel might be 1mm on
a side). The contrast to conventional photography is dramatic, since a pixel
could contain a portion of galaxy, or a part of a mouse's whisker.  This
regularity should result in better result from medical EBSR.

Finally, it is often the case that in medical applications multiple imaging
modalities are available. A patient may undergo an MRI and a CT, and with
proper registration this significantly increases the amount of information
available for EBSR.

\section{X-ray voxelization}

The goal of what I call ``x-ray voxelization'' (XRV) is to approximate the
result of a CT scan based on an x-ray, or perhaps a small number of
x-rays.\footnote{The term ``voxelization'' comes from the graphics community
  \cite{Dong:PCCG04}, where it has a distinct but related meaning.}
Technically this is a fairly straightforward application of EBSR; the main
difference is that while in conventional photography EBSR might replace an
observed low-resolution pixel by a high resolution 2x2 patch, in XRV we
replace an observed x-ray pixel by a 4x1 ``stack'' of CT voxels.

A typical CT consists of 512x512 axial slices, where by convention the x-axis
points along the patient's left-right, and the y-axis points into the patient
(assuming the patient is lying on their back in the scanner). Some number of
slices are obtained for varying choices of z, pointing from the patient's feet
to head. Depending on the exact configuration, the number of z-slices can run
into the dozens, and there can be overlap between the slices. The key here is
the y-axis, which is where the ``stack'' of reconstructed voxels will be oriented.

To be more precise, consider an x-ray taken of a patient before undergoing a
CT. (This in fact happens as part of the standard CT process, where a 'scout'
image is take by the CT scanner and used to localize the area where the full
high-resolution CT will be taken. The scout is essentially a conventional
x-ray image, though its spatial resolution is typically lower.) An individual
pixel in the scout is the average over the y-axis of the 512 voxels ``behind''
that pixel.

In x-ray voxelization the goal is to create a set of y-slices from an
x-ray. (To visualize this, imagine a standard chest x-ray, where we use EBSR
to create images at different values of y that we can navigate through. The
average of the different y-slices will be the original x-ray.) While the
technology proposed is related to tomographic imaging \cite{Kak88}, a modality
that was the direct predecessor of CT, the underlying algorithms and
techniques are quite different, especially in terms of the use of machine
learning and related inference methods.

\subsection{Theme and variations}

There are a number of straightforward extensions of the basic XRV scheme. For
example, in many situations there are several x-rays taken (chest x-rays are a
good example, as are hand x-rays). With careful registration these images can
also be used to increase the accuracy of the voxelization.

Another natural extension, suggested by Bill Freeman, is to present the user
with multiple alternative voxelizations that are supported by the input
data. Since voxelization is an ill-posed problem, at the end we have some
highly complex probability distribution over voxelizations, and we could for
example present local modes in descending order of probability.

\section{XRV research program and applications}

There are several key steps to a research program investigating XRV, some of
which can be carried out in parallel.

Despite the obvious appeal of XRV, it is necessary to identify a few clinical
applications where it would be of most use. The key properties of such an
application are that it should be common, and that a low degree of
reconstruction would be useful. It is also desirable that CT and x-rays are
commonly done, to facilitate the collection of training data.  

The natural way to proceed is to create synthetic voxelizations, where we take
a CT scan and reduce the resolution in y.  This emulates the result of perfect
noise-free voxelization at various resolutions.  We can present patients whose
diagnosis is clear on CT, and determine at what voxelization resolution the
diagnosis becomes obvious. This would effectively give us a lower bound on the
resolution that we need to achieve.

CT scouts are the next step in the process. Scouts are widely available, and
we can look at the XRV computed using the scout as the input x-ray. This is
probably the best way to investigate the various inference algorithms that
would be needed for XRV. In addition, XRV from a scout image would be of some
clinical benefit, though not on the scale of XRV from x-rays. For example, it
could indicate that a full CT is not required.

The final step in the process is to go to complete XRV, starting with an
x-ray. The scout can serve as a useful intermediate datapoint; for example, we
will need to register the x-ray to the scout. Multiple x-rays pose a more
interesting technical problem, since they need to be registered explicitly to
the 3d volume that we wish to reconstruct. Note that in general x-ray
resolution is significantly higher than CT, typically around 2k by 2k.

It is an under-appreciated fact among clinicians that CT reconstructions
themselves are not veridical, but instead rely on computational methods to
regularize an ill-posed problem.  While the CT scanner manufacturers are
closed-mouthed about their exact reconstruction methods, a review of the
literature suggests that Tikhonov regularization or its variants are likely to
be used. A natural objection to XRV is that bias in the training set would
reduce its accuracy -- for example, the training set may never contain any
images of a suitably rare condition, so how would the system properly infer
the relevant anatomy? But conventional CT has its own biases, they are just
procedural in nature as well as undocumented. Just as with CT, clinical
acceptance of XRV would follow the time-honored path of randomized clinical
trials and imilar methods to measure its diagnostic accuracy. For example, a
clinical trial could focus on patients who obtained both chest x-rays and
chest CT (which is fairly common), perhaps using biopsy results to establish
ground truth. The hope would be that for some clinically important conditions
the accuracy of XRV would be much higher than that of simple x-rays, and might
provide sufficient information to remove the need for CT in many patients.

\subsection{Applications of XRV}

XRV would have significant clinical utility, depending on the voxelization
resolution that can be computed with reasonable accuracy. X-ray machines have
near universal availability and the radiation risk is generally viewed as
quite small. The payoff for XRV would be especially large in developing
nations, where an x-ray machine and an Internet connection could give patients
some of the power of a CT scanner. But even in the US, there are plenty of
situations where CT is not immediately available, or even where XRV could be
used to determine whether or not a full CT is warranted.

\bibliographystyle{plain}
\bibliography{ALL.2015}
\end{document}